\documentclass[11pt,a4paper]{article}
\PassOptionsToPackage{breaklinks}{hyperref}
\usepackage[hyperref]{acl2019}
\usepackage[utf8]{inputenc}

\usepackage{times}
\usepackage{latexsym}
\usepackage{microtype}
\usepackage{url}
\usepackage{amsmath}
\usepackage{amssymb}
\usepackage{xcolor}
\usepackage{booktabs}
\usepackage{comment}
\usepackage{color}

\aclfinalcopy 

\usepackage{dcolumn}
\newcolumntype{d}[1]{D{.}{.}{#1}}

\title{Are We Consistently Biased? \\ Multidimensional Analysis of Biases in Distributional Word Vectors}

  \author{Anne Lauscher \and Goran Glava\v{s}\\
Data and Web Science Research Group\\ 
University of Mannheim\\ Mannheim, Germany\\
\texttt{\{anne, goran\}@informatik.uni-mannheim.de} }

\date{}

\begin{document}
\maketitle
\begin{abstract}
Word embeddings have recently been shown to reflect many of the pronounced societal biases (e.g., gender bias or racial bias). Existing studies are, however, limited in scope and do not investigate the consistency of biases across relevant dimensions like embedding models, types of texts, and different languages. In this work, we present a systematic study of biases encoded in distributional word vector spaces: we analyze how consistent the bias effects are across languages, corpora, and embedding models. Furthermore, we analyze the cross-lingual biases encoded in bilingual embedding spaces, indicative of the effects of bias transfer encompassed in cross-lingual transfer of NLP models. Our study yields some unexpected findings, e.g., that biases can be emphasized or downplayed by different embedding models or that user-generated content may be less biased than encyclopedic text. We hope our work catalyzes bias research in NLP and informs the development of bias reduction techniques. 
\end{abstract}

\section{Introduction}

%
Recent work demonstrated that word embeddings induced from large text collections encode many human biases \citep[e.g.,][]{Bolukbasi:2016:MCP:3157382.3157584,Caliskan183}. This finding is not particularly surprising given that (1) we are likely project our biases in the text that we produce and (2) these biases in text are bound to be encoded in word vectors due to the distributional nature \cite{Harris:1954} of the word embedding models \cite{mikolov2013exploiting,pennington2014glove,Bojanowski:2017tacl}. 
For illustration, consider the famous analogy-based gender bias example from \citet{Bolukbasi:2016:MCP:3157382.3157584}: \noindent\emph{``Man is to computer programmer as woman is to homemaker"}. 
This bias will be reflected in the text (i.e., the word \textit{man} will co-occur more often with words like \textit{programmer} or \textit{engineer}, whereas \textit{woman} will more often appear next to \textit{homemaker} or \textit{nurse}), and will, in turn, be captured by word embeddings built from such biased texts. While biases encoded in word embeddings can be a useful data source for diachronic analyses of societal biases \citep[e.g.,][]{GargE3635}, they may cause ethical problems for many downstream applications and NLP models. 


In order to measure the extent to which various societal biases are captured by word embeddings, \citet{Caliskan183} proposed the \emph{Word Embedding Association Test} (WEAT).
WEAT measures semantic similarity, computed through word embeddings, between two sets of \textit{target} words (e.g., insects vs. flowers) and two sets of \textit{attribute} words (e.g., pleasant vs. unpleasant words). While they test a number of biases, the analysis is limited in scope to English as the only language, GloVe \cite{pennington2014glove} as the embedding model, and Common Crawl as the type of text.
Following the same methodology, \citet{mccurdy2018} extend the analysis to three more languages (German, Dutch, Spanish), but test only for gender bias.

In this work, we present the most comprehensive study of biases captured by distributional word vector to date. We create XWEAT, a collection of multilingual and cross-lingual versions of the WEAT dataset, by translating WEAT to six other languages and offer a comparative analysis of biases over seven diverse languages. Furthermore, we measure the consistency of WEAT biases across different embedding models and types of corpora. What is more, given the recent surge of models for inducing cross-lingual embedding spaces \cite[\textit{inter alia}]{mikolov2013exploiting,hermann2014multilingual,smith2017offline,conneau2018word,artetxe2018robust,hoshen2018nonadversarial} and their ubiquitous application in cross-lingual transfer of NLP models for downstream tasks, we investigate cross-lingual biases encoded in cross-lingual embedding spaces and compare them to bias effects present of corresponding monolingual embeddings. 

Our analysis yields some interesting findings: biases do depend on the embedding model and, quite surprisingly, they seem to be less pronounced in embeddings trained on social media texts. Furthermore, we find that the effects (i.e., amount) of bias in cross-lingual embedding spaces can roughly be predicted from the bias effects of the corresponding monolingual embedding spaces.

\section{Data for Measuring Biases}
We first introduce the WEAT dataset \cite{Caliskan183} and then describe XWEAT, our multilingual and cross-lingual extension of WEAT designed for comparative bias analyses across languages and in cross-lingual embedding spaces. 

\subsection{WEAT}
\label{sec:data}

The Word Embedding Association Test (WEAT) \cite{Caliskan183} is an adaptation of the Implicit Association Test (IAT) \cite{Nosek02harvestingimplicit}. Whereas IAT measures biases based on response times of human subjects to provided stimuli, WEAT quantifies the biases using semantic similarities between word embeddings of the same stimuli. For each bias test, WEAT specifies four stimuli sets: two sets of \textit{target} words and two sets of \textit{attribute} words.  %
%
The sets of target words represent stimuli \emph{between} which we want to measure the bias (e.g., for gender biases, one target set could contain male names and the other females names). The \textit{attribute} words, on the other hand, represent stimuli \emph{towards} which the bias should be measured (e.g., one list could contain pleasant stimuli like \textit{health} and \textit{love} and the other negative \textit{war} and \textit{death}). The WEAT dataset defines ten bias tests, each containing two target and two attribute sets.\footnote{Some of the target and attribute sets are shared across multiple tests.} Table \ref{tbl:weat} enumerates the WEAT tests and provides examples of the respective target and attribute words.

\setlength{\tabcolsep}{3pt}
\begin{table*}[t]
\centering
{\fontsize{8pt}{8pt}\selectfont
\begin{tabular}{c | llll}
\toprule
\textbf{Test} & \textbf{Target Set \#1} & \textbf{Target Set \#2} & \textbf{Attribute Set \#1} & \textbf{Attribute Set \#2} \\ \midrule
T1 & Flowers (e.g.,~\emph{aster}, \emph{tulip})
& Insects (e.g.,~\emph{ant}, \emph{flea}) & Pleasant (e.g.,~\emph{health}, \emph{love}) & Unpleasant~(e.g., \emph{abuse})
\\
T2 &  Instruments (e.g.,~\emph{cello}, \emph{guitar}) 
& Weapons (e.g.,~\emph{gun}, \emph{sword}) &
Pleasant & Unpleasant\\
T3 & Euro-American names (e.g.,~\emph{Adam})
& Afro-American names (e.g.,~\emph{Jamel})
& Pleasant (e.g.,~\emph{caress}) & Unpleasant (e.g.,~\emph{abuse})\\
T4 & Euro-American names (e.g.,~\emph{Brad})
& Afro-American names (e.g.,~\emph{Hakim})
& Pleasant 
& Unpleasant 
\\
T5 & Euro-American names
& Afro-American names
& Pleasant  (e.g.,~\emph{joy}) & Unpleasant  (e.g.,~\emph{agony})\\
T6 & Male names (e.g.,~\emph{John}) & Female names (e.g.,~\emph{Lisa}) & Career (e.g. \emph{management})
& Family (e.g.,~\emph{children}) \\
T7 & Math (e.g.,~\emph{algebra}, \emph{geometry}) & Arts (e.g.,~\emph{poetry}, \emph{dance}) & Male (e.g.,~\emph{brother}, \emph{son}) & Female (e.g.,~\emph{woman}, \emph{sister}) \\
T8 & Science (e.g.,~\emph{experiment}) & Arts & Male & Female\\
T9 & Physical condition (e.g.,~\emph{virus}) & Mental condition (e.g.,~\emph{sad}) & Long-term (e.g.,~\emph{always}) & Short-term (e.g.,~\emph{occasional})\\
T10 & Older names (e.g.,~\emph{Gertrude}) & Younger names (e.g.,~\emph{Michelle}) & Pleasant & Unpleasant \\ \bottomrule
\end{tabular}
}
\caption{WEAT bias tests.}
\label{tbl:weat}
\end{table*}

\subsection{Multilingual and Cross-Lingual WEAT}
\label{sec:xweat}

We port the WEAT tests to the multilingual and cross-lingual settings by translating the test vocabularies consisting of attribute and target terms from English to six other languages: German (\textsc{de}), Spanish (\textsc{es}), Italian (\textsc{it}), Russian (\textsc{ru}), Croatian (\textsc{hr}), and Turkish (\textsc{tr}). We first automatically translate the vocabularies and then let native speakers of the respective languages (also fluent in English) fix the incorrect automatic translations (or introduce better fitting ones). Our aim was to translate WEAT vocabularies to languages from diverse language families\footnote{English and German from the Germanic branch of Indo-European languages, Italian and Spanish from the Romance branch, Russian and Croatian from the Slavic branch, and finally Turkish as a non-Indo-European language.} for which we also had access to native speakers. Whenever the translation of an English term indicated the gender in a target language (e.g., \emph{Freund} vs. \emph{Freundin} in \textsc{de}), we asked the translator to provide both male and female forms and included both forms in the respective test vocabularies. This helps avoiding artificially amplifying the gender bias stemming from the grammatically masculine or feminine word forms.

The monolingual tests in other languages are created by simply using the corresponding translations of target and attribute sets in those languages. For every two languages, L1 and L2 (e.g., \textsc{de} and \textsc{it}), we create two cross-lingual bias tests: we pair (1) target translations in L1 with L2 translations of attributes (e.g., for T2 we combine \textsc{de} target sets \{\textit{Klavier}, \textit{Cello}, \textit{Gitarre}, \dots\} and \{\textit{Gewehr}, \textit{Schwert}, \textit{Schleuder}, \dots\} with \textsc{it} attribute sets \{\textit{salute}, \textit{amore}, \textit{pace}, \dots\} and \{\textit{abuso}, \textit{omicidio}, \textit{tragedia}, \dots\}), and vice versa, (2) target translations in L2 with attribute translations in L1 (e.g., for T2, \textsc{it} target sets \{\textit{pianoforte}, \textit{violoncello}, \textit{chitarra}, \dots\} and \{\textit{fucile}, \textit{spada}, \textit{fionda}, \dots\} with \textsc{de} attribute sets \{\textit{Gesundheit}, \textit{Liebe}, \textit{Frieden}, \dots\} and \{\textit{Missbrauch}, \textit{Mord}, \textit{Tragödie}, \dots\}). We did not translate or modify proper names from WEAT sets 3--6. In our multilingual and cross-lingual experiments we, however, discard the (translations of) WEAT tests for which we cannot find more than 20\% of words from some target or attribute set in the embedding vocabulary of the respective language. This strategy eliminates tests 3--5 and 10 which include proper American names, majority of which can not be found in distributional vocabularies of other languages. The exception to this is test 6, containing frequent English first names (e.g., \textit{Paul}, \textit{Lisa}), which we do find in distributional vocabularies of other languages as well. In summary, for languages other than \textsc{en} and for cross-lingual settings, we execute six bias tests (T1, T2, T6--T9).  



\section{Methodology}
We adopt the general bias-testing framework from \newcite{Caliskan183}, but we span our study over multiple dimensions: (1) corpora -- we analyze the consistency of biases across distributional vectors induced from different types of text; (2) embedding models -- we compare biases across distributional vectors induced by different embedding models (on the same corpora); and (3) languages -- we measure biases for word embeddings of different languages, trained from comparable corpora. Furthermore, unlike \newcite{Caliskan183}, we test whether biases depend on the selection of the similarity metric. Finally, given the ubiquitous adoption of cross-lingual embeddings \cite{ruder2017survey,glavas2019properly}, we investigate biases in a variety of bilingual embedding spaces.      

\vspace{1.4mm}

\noindent\textbf{Bias-Testing Framework.} We first describe the WEAT framework \cite{Caliskan183}. 
Let $X$ and $Y$ be two sets of \textit{targets}, and $A$ and $B$ two sets of \textit{attributes} (see \S \ref{sec:data}). The tested statistic is the difference between $X$ and $Y$ in average similarity of their terms with terms from $A$ and $B$: \vspace{-1em}

{\small
\begin{equation}
    s(X, Y, A, B) = \sum_{x \in X}{s(x, A, B)} - \sum_{y \in Y}{s(y, A, B)}\,, 
\end{equation}}

\vspace{-1em}

\noindent with association difference for term $t$ computed as: 
\vspace{-2.5em}

{\small
\begin{equation}
    s(t, A, B) = \frac{1}{|A|}\sum_{a \in A}{f(\mathbf{t}, \mathbf{a})} - \frac{1}{|B|}\sum_{b \in B}{f(\mathbf{t}, \mathbf{b})}\,, 
\end{equation}}

\vspace{-1em}

\noindent where $\mathbf{t}$ is the distributional vector of term $t$ and $f$ is a similarity or distance metric, fixed to cosine similarity in the original work \cite{Caliskan183}. The significance of the statistic is validated by comparing the score $s(X, Y, A, B)$ with the scores $s(X_i, Y_i, A, B)$ obtained for different equally sized partitions $\{X_i, Y_i\}_i$ of the set $X \cup Y$. The $p$-value of this permutation test is then measured as the probability of $s(X_i, Y_i, A, B) > s(X, Y, A, B)$ computed over all permutations $\{X_i, Y_i\}_i$.\footnote{If $f$ is a distance rather than a similarity metric, we measure the probability of $s(X_i, Y_i, A, B) < s(X, Y, A, B)$.} The effect size, that is, the ``amount of bias'', is computed as the normalized measure of separation between association distributions: \vspace{-1em} 

{\small
\begin{equation}
\frac{\mu\left(\{s(x, A, B)\}_{x \in X}\right) - \mu\left(\{s(y, A, B)\}_{y \in Y}\right)}{\sigma\left(\{s(w, A, B)\}_{w \in X \cup Y}\right)}\,,
\end{equation}}

\vspace{-1em}
\noindent where $\mu$ denotes the mean and $\sigma$ standard deviation. 

\vspace{1.4mm}

\noindent \textbf{Dimensions of Bias Analysis.} We analyze the bias effects across multiple dimensions. First, we analyze the effect that different embedding models have: we compare biases of distributional spaces induced from English Wikipedia, using CBOW \cite{mikolov2013distributed}, \textsc{GloVe} \cite{pennington2014glove}, \textsc{fastText} \cite{Bojanowski:2017tacl}, and \textsc{Dict2Vec} algorithms \cite{tissier2017dict2vec}. 
Secondly, we investigate the effects of biases in different corpora: we compare biases between embeddings trained on the Common Crawl, Wikipedia, and a corpus of tweets.  
Finally, and (arguably) most interestingly, we test the consistency of biases across seven languages (see \S\ref{sec:xweat}). To this end, we test for biases in seven monolingual \textsc{fastText} spaces trained on Wikipedia dumps of the respective languages. \vspace{1.4mm}

\noindent \textbf{Biases in Cross-Lingual Embeddings.} Cross-lingual embeddings (CLEs) are widely used in multilingual NLP and cross-lingual transfer of NLP models. Despite the ubiquitous usage of CLEs, the biases they potentially encode have not been analyzed so far. 
%
%
We analyze projection-based CLEs \cite{glavas2019properly}, induced through post-hoc linear projections between monolingual embedding spaces \cite{mikolov2013exploiting,artetxe2016learning,smith2017offline}. The projection is commonly learned through supervision with few thousand word translation pairs. Most recently, however, a number of models have been proposed that learn the projection without any bilingual signal \cite[\textit{inter alia}]{artetxe2018robust,conneau2018word,hoshen2018nonadversarial,alvarez2018gromov}. 
Let $\mathbf{X}$ and $\mathbf{Y}$ be, respectively, the distributional spaces of the source (S) and target (T) language and let $D = \{w^i_{S}, w^i_{T}\}_i$ be the word translation dictionary. Let $(\mathbf{X}_S, \mathbf{X}_T)$ be the aligned subsets of monolingual embeddings, corresponding to word-aligned pairs from $D$. We then compute the orthogonal matrix $\mathbf{W}$ that minimizes the Euclidean distance between $\mathbf{X_S}\mathbf{W}$ and $\mathbf{X_T}$ \cite{smith2017offline}: 
%
$\mathbf{W} = \mathbf{UV}^\top$,
%
 where $\mathbf{U\Sigma V}^\top = \mathit{SVD}(\mathbf{X}_{T} {\mathbf{X}_{S}}^\top)$. We create comparable bilingual dictionaries $D$ by translating 5K most frequent \textsc{en} words to other six languages and induce a bilingual space for all $21$ language pairs.   


\section{Findings}
Here, we report and discuss the results of our multi-dimensional analysis.
Table \ref{tbl:measure} shows the effect sizes for WEAT T1--T10 based on Euclidean or cosine similarity between word vector representations trained on the \textsc{en} Wikipedia using \textsc{fastText}.
\setlength{\tabcolsep}{2.4pt}
\begin{table}[t]
\centering
\small{
\begin{tabular}{l cccccccccc}
\toprule
\textbf{Metric} & T1 & T2 & T3 & T4 & T5 & T6 & T7 & T8 & T9 & T10 \\ \midrule
\textbf{Cos} & 1.7 & 1.6 & -0.1$^*$ & -0.2$^*$ & -0.2$^*$ & 1.8 & 1.3 & 1.3 & 1.7 & -0.6$^*$ \\ 
\textbf{Euc} & 1.7 & 1.6 & -0.1$^*$ & -0.2$^*$ & -0.1$^*$ & 1.8 & 1.3 & 1.3 & 1.7 & -0.7$^*$ \\ \bottomrule
\end{tabular}
}
\caption{WEAT bias effects (\textsc{en} \textsc{fastText} embeddings trained on Wikipedia) for cosine similarity and Euclidean distance. Asterisks indicate bias effects that are insignificant at $\alpha < 0.05$.}
\label{tbl:measure}
\vspace{-1em}
\end{table}
We observe the highest bias effects for T6 (Male/Female -- Career/Family), T9 (Physical/Mental deseases -- Long-term/Short-term), and T1 (Insects/Flowera -- Positive/Negative). 
Importantly, the results show that biases do not depend on the similarity metric. We observe nearly identical effects for cosine similarity and Euclidean distance for all WEAT tests. In the following experiments we thus analyze biases only for cosine similarity. 
%

\vspace{1.4mm}

\noindent \textbf{Word Embedding Models.} Table \ref{tbl:en_methods} compares biases in embedding spaces induced with different models: \textsc{CBOW}, \textsc{GloVe}, \textsc{fastText}, and \textsc{Dict2Vec}. While the first three embedding methods are trained on Wikipedia only, \textsc{Dict2Vec} employs definitions from dictionaries (e.g., Oxford dictionary) as additional resources for identifying strongly related terms.\footnote{Two terms A and B are strongly related if B appears in the definition of A and vice versa \citep{tissier2017dict2vec}.} We only report WEAT test results T1, T2, and T7--T9 for \textsc{Dict2Vec}, as the \textsc{Dict2Vec}'s vocabulary does not cover most of the proper names from the remaining tests.
%

\setlength{\tabcolsep}{1.8pt}
\begin{table}[t]
\centering
\small{
\begin{tabular}{l d{2.2} d{2.2} d{2.2} d{2.2}}
\toprule
WEAT  & \multicolumn{1}{c}{\textbf{\textsc{CBOW}}} & \multicolumn{1}{c}{\textbf{\textsc{GloVe}}} & \multicolumn{1}{c}{\textbf{\textsc{FastText}}} & \multicolumn{1}{c}{\textbf{\textsc{Dict2Vec}}} \\ \midrule
  T1 & 1.20 & 1.41 & 1.67 & 1.35 \\
  T2 & 1.38 & 1.45 & 1.55 & 1.66 \\
  T3 & -0.28^{*} & 1.16 & -0.09^{*} & \multicolumn{1}{c}{--} \\
  T4 & -0.35^{*} & 1.36 & -0.17^{*} & \multicolumn{1}{c}{--} \\
  T5 & -0.36^{*} & 1.40 & -0.18^{*} & \multicolumn{1}{c}{--} \\
  T6 & 1.78 & 1.75 & 1.83 & \multicolumn{1}{c}{--} \\
  T7 & 1.28 & 1.16 & 1.30 & 1.48 \\
  T8 & 0.39^{*} & 1.28^{*} & 1.30 & 1.30 \\
  T9 & 1.55 & 1.35 & 1.72 & 1.69 \\
  T10 & 0.09^{*} & 1.17 & -0.61^{*} & \multicolumn{1}{c}{--} \\
\bottomrule
\end{tabular}
}
\caption{WEAT bias effects for spaces induced (on \textsc{en} Wikipedia) with different embedding models: \textsc{CBOW}, \textsc{GloVe}, \textsc{fastText}, and \textsc{Dict2Vec} methods. Asterisks indicate bias effects that are insignificant at $\alpha < 0.05$.} 
\label{tbl:en_methods}
\end{table}

Somewhat surprisingly, the bias effects seem to vary greatly across embedding models. While \textsc{GloVe} embeddings are biased according to all tests,\footnote{This is consistent with the original results obtained by \newcite{Caliskan183}.} \textsc{fastText} and especially \textsc{CBOW} exhibit significant biases only for a subset of tests. We hypothesize that the bias effects reflected in the distributional space depend on the preprocessing steps of the embedding model. \textsc{fastText}, for instance, relies on embedding subword information, in order to avoid issues with representations of out-of-vocabulary and underrepresented terms: additional reliance on morpho-syntactic signal may make \textsc{fastText} more resilient to biases stemming from distributional signal (i.e., word co-occurrences). The fact that the embedding space induced with \textsc{Dict2Vec} exhibits larger bias effects may seem counterintuitive at first, since the dictionaries used for vector training should be more objective and therefore less biased than encyclopedic text. We believe, however, that the additional dictionary-based training objective only propagates the distributional biases across definitionally related words. Generally, we find these results to be important as they indicate that embedding models may accentuate or diminish biases expressed in text.
\vspace{1.4mm}


\noindent\textbf{Corpora.} In Table \ref{tbl:en_corpora} we compare the biases of embeddings trained with the same model (\textsc{GloVe}) but on different corpora: Common Crawl (i.e., noisy web content), Wikipedia (i.e., encyclopedic text) and a corpus of tweets (i.e., user-generated content).
\setlength{\tabcolsep}{2.5pt}
\begin{table}[t]
\centering
\small{
\begin{tabular}{l d{1.1}d{1.1}d{1.1}d{1.1}d{1.1}d{1.1}d{2.1}d{1.1}d{1.1}d{1.1}}
\toprule
\textbf{Corpus}  & \multicolumn{1}{c}{T1} & \multicolumn{1}{c}{T2} & \multicolumn{1}{c}{T3} & \multicolumn{1}{c}{T4} & \multicolumn{1}{c}{T5} & \multicolumn{1}{c}{T6} & \multicolumn{1}{c}{T7} & \multicolumn{1}{c}{T8} & \multicolumn{1}{c}{T9} & \multicolumn{1}{c}{T10} \\ \midrule
\textbf{\textsc{Wiki}}   & 1.4 & 1.5 & 1.2 & 1.4 & 1.4 & 1.8 & 1.2 & 1.3 & 1.3 & 1.2 \\ 
\textbf{\textsc{CC}}  & 1.5 & 1.6 & 1.5 & 1.6 & 1.4 & 1.9 & 1.1 & 1.3 & 1.4 & 1.3  \\ 
\textbf{\textsc{Tweets}} & 1.2 & 1.0 & 1.1 & 1.2 & 1.2 & 1.2 & -0.2^* & 0.6^* & 0.7^* & 0.8^*  \\ 
\bottomrule
\end{tabular}
}
\caption{WEAT bias effects for \textsc{GloVe} embeddings trained on different corpora: Wikipedia (\textsc{Wiki}), Common Crawl (\textsc{CC}), and corpus of tweets (\textsc{Tweets}). Asterisks indicate bias effects that are insignificant at $\alpha < 0.05$.}
\label{tbl:en_corpora}
\vspace{-1.5em}
\end{table}
\setlength{\tabcolsep}{3.0pt}
\begin{table}[t!]
\centering
\small{
\begin{tabular}{l d{2.2} d{2.2} d{2.2} d{2.2} d{2.2} d{2.2} d{2.2}}
\toprule
XW & \multicolumn{1}{c}{\textbf{\textsc{en}}} & \multicolumn{1}{c}{\textbf{\textsc{de}}} & \multicolumn{1}{c}{\textbf{\textsc{es}}} & \multicolumn{1}{c}{\textbf{\textsc{it}}} & \multicolumn{1}{c}{\textbf{\textsc{hr}}} & \multicolumn{1}{c}{\textbf{\textsc{ru}}} & \multicolumn{1}{c}{\textbf{\textsc{tr}}} \\ 
\midrule
T1 & 1.67 & 1.36 & 1.47 & 1.28 & 1.45 & 1.28 & 1.21 \\
T2 & 1.55 & 1.25 & 1.47 & 1.36 & 1.10 & 1.46 & 0.83 \\
T6 & 1.83 & 1.59 & 1.67 &  1.72 & 1.83 & 1.87 & 1.85 \\
T7 & 1.30 & 0.46^* & 1.47 & 1.00 & 0.72^* & 0.59^* & -0.88 \\
T8 & 1.30 & 0.05^* & 1.16 & 0.10^* & 0.13^* & 0.37^* & 1.72 \\
T9 & 1.72 & 0.82^* & 1.71 & 1.57 & -0.40^* & 1.73 & 1.09^* \\
\midrule
$\mathit{Avg}_{\mathit{all}}$ & 1.56 & 0.92 & 1.49 & 1.17 & 0.81 & 1.22 & 0.88 \\
$\mathit{Avg}_{\mathit{sig}}$ & 1.68 & 1.4 & 1.54 & 1.45 & 1.46 & 1.54 & 1.30 \\ \bottomrule
\end{tabular}
}
\caption{XWEAT effects across languages (\textsc{fastText} embeddings trained on Wikipedias). $\mathit{Avg}_{\mathit{all}}$: average of effects over all tests; $\mathit{Avg}_{\mathit{sig}}$: average over the subset of tests yielding significant biases for all languages. Asterisks indicate bias effects that are insignificant at $\alpha < 0.05$.}
\label{tbl:multiling_wiki}
\end{table}
\setlength{\tabcolsep}{3.5pt}
\begin{table}[ht!]
\centering
\small{
\begin{tabular}{c | d{2.2} d{2.2} d{2.2} d{2.2} d{2.2} d{2.2} d{2.2}}
\toprule
 XW & \multicolumn{1}{c}{\textbf{\textsc{en}}} & \multicolumn{1}{c}{\textbf{\textsc{de}}} & \multicolumn{1}{c}{\textbf{\textsc{es}}} & \multicolumn{1}{c}{\textbf{\textsc{it}}} & \multicolumn{1}{c}{\textbf{\textsc{hr}}} & \multicolumn{1}{c}{\textbf{\textsc{ru}}} & \multicolumn{1}{c}{\textbf{\textsc{tr}}} \\  \midrule
\textbf{\textsc{en}} & \multicolumn{1}{c}{--} &1.09^*&1.58&1.49&0.72^*&1.17^*&1.20^*\\
\textbf{\textsc{de}} &1.53& \multicolumn{1}{c}{--} &1.50&1.45&0.55^*&1.35&1.07^*\\
\textbf{\textsc{es}} &1.52&0.79^*& \multicolumn{1}{c}{--} &1.38^*&0.60^*&1.37^*&1.09^*\\
\textbf{\textsc{it}} &1.33^*&0.69^*&1.27& \multicolumn{1}{c}{--} &0.53^*&0.82^*&0.80^*\\
\textbf{\textsc{hr}} &1.47&1.30^*&1.29&1.18^*& \multicolumn{1}{c}{--} &1.14^*&1.11^*\\
\textbf{\textsc{ru}} &1.47&0.72^*&1.35&1.35&0.77^*& \multicolumn{1}{c}{--} &0.80^*\\
\textbf{\textsc{tr}} &1.41&0.90^*&1.37^*&1.45&0.29^*&0.64^*& \multicolumn{1}{c}{--} \\
\bottomrule
\end{tabular}
}
\caption{XWEAT bias effects (aggregated over all six tests) for cross-lingual word embedding spaces. Rows: \textit{targets} language; columns: \textit{attributes} language. Asterisks indicate the inclusion of bias effects sizes in the aggregation that were insignificant at $\alpha < 0.05$.}
\label{tbl:xling_wiki}
\end{table}
\noindent Expectedly, the biases are slightly more pronounced for embeddings trained on the Common Crawl than for those obtained on encyclopedic texts (Wikipedia). Countering our intuition, the corpus of tweets seems to be consistently less biased (across all tests) than Wikipedia. In fact, the biases covered by tests T7--T10 are not even significantly present in the vectors trained on tweets. This finding is indeed surprising and the phenomenon warrants further investigation. 
\vspace{1.4mm}

\noindent \textbf{Multilingual Comparison.} Table \ref{tbl:multiling_wiki} compares the bias effects across the seven different languages. 
%
%
Whereas many of the biases are significant in all languages, \textsc{de}, \textsc{hr}, and \textsc{tr} consistently display smaller effect sizes. Intuitively, the amount of bias should be proportional to the size of the corpus.\footnote{The larger the corpus the larger is the overall number of contexts in which some bias may be expressed.} Wikipedias in \textsc{tr} and \textsc{hr} are the two smallest ones -- thus they are expected to contain least biased statements. \textsc{de} Wikipedia, on the other hand, is the second largest and low bias effects here suggest that German texts are indeed less biased than texts in other languages.
%
Additionally, for (X)WEAT T2, which defines a universally accepted bias (Instruments vs. Weapons), \textsc{tr} and \textsc{hr} exhibit the smallest effect sizes, while the highest bias is observed for \textsc{en} and \textsc{it}. We measure the highest gender bias, according to (X)WEAT T6, for \textsc{tr} and \textsc{ru}, and the lowest for \textsc{de}. 

\vspace{1.4mm} 

\noindent \textbf{Biases in Cross-Lingual Embeddings}. We report bias effects for all 21 bilingual embedding spaces in Table \ref{tbl:xling_wiki}. For brevity, here we report the bias effects averaged over all six XWEAT tests (we provide results detailing bias effects for each of the tests separately in the supplementary materials). Generally, the bias effects of bilingual spaces are in between the bias effects of the two corresponding monolingual spaces (cf.~Table \ref{tbl:multiling_wiki}): this means that we can roughly predict the amount of bias in a cross-lingual embedding space from the same bias effects of corresponding monolingual spaces. For example, effects in cross-lingual spaces increase over monolingual effects for low-bias languages (\textsc{hr} and \textsc{tr}), and decrease for high-bias languages (\textsc{en} and \textsc{es}). 


\section{Conclusion}
In this paper, we have presented the largest study to date on biases encoded in distributional word vector spaces. To this end, we have extended previous analyses based on the WEAT test \cite{Caliskan183,mccurdy2018} in multiple dimensions: across seven languages, four embedding models, and three different types of text. We find that different models may produce embeddings with very different biases, which stresses the importance of embedding model selection when fair text representations are to be created. Surprisingly, we find that the user-generated texts, such as tweets, may be less biased than redacted content. Furthermore, we have investigated the bias effects in cross-lingual embedding spaces and have shown that they may be predicted from the biases of corresponding monolingual embeddings. We make the XWEAT dataset and the testing code publicly available,\footnote{At: \url{https://github.com/umanlp/XWEAT}.} hoping to fuel further research on biases encoded in word representations. 

\section*{Acknowledgments}
We thank our six native speakers for manually correcting and improving (i.e., post-editing) the automatic translations of the WEAT dataset. 

\bibliography{acl2019}
\bibliographystyle{acl_natbib}

\newpage
\appendix
\section{Additional Results}
For completeness, we report detailed results on bias effects for each of the six XWEAT tests and bilingual word embedding spaces for all 21 language pairs. Tables \ref{tbl:xling_wiki_1} to \ref{tbl:xling_wiki_9} show bias effects for XWEAT tests T1, T2, and T6--T9.

\setlength{\tabcolsep}{5pt}
\begin{table}[t]
\centering
\small{
\begin{tabular}{c | d{1.2} d{1.2} d{1.2} d{1.2} d{1.2} d{1.2} d{1.2}}
\toprule
XW1 & \multicolumn{1}{c}{\textbf{\textsc{en}}} & \multicolumn{1}{c}{\textbf{\textsc{de}}} & \multicolumn{1}{c}{\textbf{\textsc{es}}} & \multicolumn{1}{c}{\textbf{\textsc{it}}} & \multicolumn{1}{c}{\textbf{\textsc{hr}}} & \multicolumn{1}{c}{\textbf{\textsc{ru}}} & \multicolumn{1}{c}{\textbf{\textsc{tr}}} \\ \midrule 
\textbf{\textsc{en}} & \multicolumn{1}{c}{--} & 1.28 & 1.63 & 1.62 & 1.59 & 1.49 & 1.32 \\ 
\textbf{\textsc{de}} & 1.55 & \multicolumn{1}{c}{--} & 1.28 & 1.45 & 1.41 & 1.03 & 1.29 \\ 
\textbf{\textsc{es}} & 1.45 & 1.25 & \multicolumn{1}{c}{--} & 1.28 & 1.21 & 1.31 & 1.09 \\ 
\textbf{\textsc{it}} & 1.18 & 1.10 & 1.28 & \multicolumn{1}{c}{--} & 1.29 & 0.61 & 1.09 \\ 
\textbf{\textsc{hr}} & 1.57 & 1.62 & 1.59 & 1.62 & \multicolumn{1}{c}{--} & 1.62 & 1.63 \\ 
\textbf{\textsc{ru}} & 1.41 & 1.12 & 1.20 & 1.38 & 1.46 & \multicolumn{1}{c}{--} & 1.29 \\ 
\textbf{\textsc{tr}} & 1.23 & 1.21 & 1.06 & 1.26 & 1.24 & 1.04 & \multicolumn{1}{c}{--} \\ 
\bottomrule
\end{tabular}
}
\caption{XWEAT T1 effect sizes for cross-lingual embedding spaces. Rows denote the target set language, column the attribute set language.}
\label{tbl:xling_wiki_1}
\end{table}

\setlength{\tabcolsep}{5pt}
\begin{table}[t]
\centering
\small{
\begin{tabular}{c | c c c c c c c}
\toprule
XW2  & \textbf{\textsc{en}} & \textbf{\textsc{de}} & \textbf{\textsc{es}} & \textbf{\textsc{it}} & \textbf{\textsc{hr}} & \textbf{\textsc{ru}} & \textbf{\textsc{tr}} \\  \midrule
\textbf{\textsc{en}} & -- &1.35 &1.51& 1.48 &1.60 &1.56 & 1.15\\
\textbf{\textsc{de}} &1.37 & -- &1.25&1.19&1.31&1.47&1.16\\
\textbf{\textsc{es}} &1.55 &1.50 & -- &1.53&1.50&1.57 &1.22\\
\textbf{\textsc{it}} &1.54 &1.37 &1.28& -- &1.47&1.39&1.27\\
\textbf{\textsc{hr}} &1.19 &1.25 &0.72 &1.09& -- &1.26&0.81\\
\textbf{\textsc{ru}} &1.46 &1.26 &1.23&1.08&1.13& -- &0.71\\
\textbf{\textsc{tr}} &1.29 &1.44 &1.21 &1.4&1.25&1.57& -- \\
\bottomrule
\end{tabular}
}
\caption{XWEAT T2 effect sizes for cross-lingual embedding spaces. Rows denote the target set language, column the attribute set language.}
\label{tbl:xling_wiki_2}
\end{table}

\setlength{\tabcolsep}{5pt}
\begin{table}[t]
\centering
\small{
\begin{tabular}{c | c c c c c c c}
\toprule
XW6 & \textbf{\textsc{en}} & \textbf{\textsc{de}} & \textbf{\textsc{es}} & \textbf{\textsc{it}} & \textbf{\textsc{hr}} & \textbf{\textsc{ru}} & \textbf{\textsc{tr}} \\ \midrule
\textbf{\textsc{en}} & -- & 1.77&1.81&1.88&1.83&1.78&1.89\\
\textbf{\textsc{de}} &1.82& -- &1.77&1.85&1.84&1.74&1.86\\
\textbf{\textsc{es}} &1.71&0.95& -- &1.81&1.80&1.61&1.50\\
\textbf{\textsc{it}} &1.76&1.58&1.703& -- &1.72&1.77&1.76\\
\textbf{\textsc{hr}} &1.68&1.65&1.66&1.43& -- &1.74&1.73\\
\textbf{\textsc{ru}} &1.86&1.74&1.74&1.82&1.86& -- &1.80\\
\textbf{\textsc{tr}} &1.90&1.66&1.77&1.82&1.77&1.55& -- \\
\bottomrule
\end{tabular}
}
\caption{XWEAT T6 effect sizes for cross-lingual embedding spaces. Rows denote the target set language, column the attribute set language.}
\label{tbl:xling_wiki_6}
\end{table}

\setlength{\tabcolsep}{3pt}
\begin{table}[t]
\centering
\small{
\begin{tabular}{c | d{2.2} d{2.2} d{2.2} d{2.2} d{2.2} d{2.2} d{2.2}}
\toprule
XW7 & \multicolumn{1}{c}{\textbf{\textsc{en}}} & \multicolumn{1}{c}{\textbf{\textsc{de}}} & \multicolumn{1}{c}{\textbf{\textsc{es}}} & \multicolumn{1}{c}{\textbf{\textsc{it}}} & \multicolumn{1}{c}{\textbf{\textsc{hr}}} & \multicolumn{1}{c}{\textbf{\textsc{ru}}} & \multicolumn{1}{c}{\textbf{\textsc{tr}}} \\
\midrule
\textbf{\textsc{en}} & \multicolumn{1}{c}{--} &0.34^*&1.36&1.33&0.26^*&0.46^*&0.49^*\\
\textbf{\textsc{de}} &1.51& \multicolumn{1}{c}{--} &1.60&1.42&0.23^*&1.33&-0.62^*\\
\textbf{\textsc{es}} &1.63&0.24^*& \multicolumn{1}{c}{--} &1.26&0.60^*&1.29&1.55\\
\textbf{\textsc{it}} &1.12&0.65^*&1.01& \multicolumn{1}{c}{--} &0.51^*&-0.20^*&-1.08\\
\textbf{\textsc{hr}} &1.46&0.94&0.95&1.27& \multicolumn{1}{c}{--} &0.62^*&0.00^*\\
\textbf{\textsc{ru}} &1.19&-0.51^*&1.30&1.09&0.81^*& \multicolumn{1}{c}{--} &-0.79^*\\
\textbf{\textsc{tr}} &1.22&0.07^*&0.81^*&1.30&-0.23^*&-0.48^*& \multicolumn{1}{c}{--} \\
\bottomrule
\end{tabular}
}
\caption{XWEAT T7 effect sizes for cross-lingual embedding spaces. Rows denote the target set language, column the attribute set language.}
\label{tbl:xling_wiki_7}
\end{table}

\setlength{\tabcolsep}{3pt}
\begin{table}[t]
\centering
\small{
\begin{tabular}{c | d{2.2} d{2.2} d{2.2} d{2.2} d{2.2} d{2.2} d{2.2}}
\toprule
XW8 & \multicolumn{1}{c}{\textbf{\textsc{en}}} & \multicolumn{1}{c}{\textbf{\textsc{de}}} & \multicolumn{1}{c}{\textbf{\textsc{es}}} & \multicolumn{1}{c}{\textbf{\textsc{it}}} & \multicolumn{1}{c}{\textbf{\textsc{hr}}} & \multicolumn{1}{c}{\textbf{\textsc{ru}}} & \multicolumn{1}{c}{\textbf{\textsc{tr}}} \\
\midrule
\textbf{\textsc{en}} & \multicolumn{1}{c}{--} &0.68^*&1.49&1.01&-0.38^*&-0.06^*&0.71^*\\
\textbf{\textsc{de}} &1.17 & \multicolumn{1}{c}{--} &1.43&1.10&-0.09^*&1.06&1.16\\
\textbf{\textsc{es}} &1.13&-0.69^*& \multicolumn{1}{c}{--} &0.61^*&-0.19^*&0.67^*&-0.18^*\\
\textbf{\textsc{it}} &0.75^*&-0.76^*&0.87& \multicolumn{1}{c}{--} &-0.18^*&-0.52^*&0.04^*\\
\textbf{\textsc{hr}} &1.36&0.42^*&0.92&0.76^*& \multicolumn{1}{c}{--} &-0.16^*&0.90\\
\textbf{\textsc{ru}} &1.09&-0.84^*&0.96&0.99&0.19^*& \multicolumn{1}{c}{--} &1.00\\
\textbf{\textsc{tr}} &0.93&0.06^*&1.49&1.21&-0.47^*&-0.43^*& \multicolumn{1}{c}{--} \\
\bottomrule
\end{tabular}
}
\caption{XWEAT T8 effect sizes for cross-lingual embedding spaces. Rows denote the target set language, column the attribute set language.}
\label{tbl:xling_wiki_8}
\end{table}

\setlength{\tabcolsep}{3pt}
\begin{table}[t]
\centering
\small{
\begin{tabular}{c |d{2.2} d{2.2} d{2.2} d{2.2} d{2.2} d{2.2} d{2.2}}
\toprule
XW9 & \multicolumn{1}{c}{\textbf{\textsc{en}}} & \multicolumn{1}{c}{\textbf{\textsc{de}}} & \multicolumn{1}{c}{\textbf{\textsc{es}}} & \multicolumn{1}{c}{\textbf{\textsc{it}}} & \multicolumn{1}{c}{\textbf{\textsc{hr}}} & \multicolumn{1}{c}{\textbf{\textsc{ru}}} & \multicolumn{1}{c}{\textbf{\textsc{tr}}} \\
\midrule
\textbf{\textsc{en}} & \multicolumn{1}{c}{--} &1.12&1.66&1.61&-0.59^*&1.76 &1.65\\
\textbf{\textsc{de}} &1.74& \multicolumn{1}{c}{--} &1.68&1.66&-1.39&1.46 &1.57 \\
\textbf{\textsc{es}} &1.64&1.48& \multicolumn{1}{c}{--} &1.79&-1.34&1.75&1.37 \\
\textbf{\textsc{it}} &1.62&0.19^*&1.47& \multicolumn{1}{c}{--} &-1.63&1.87&1.74\\
\textbf{\textsc{hr}} &1.54&1.89&1.87&0.96^*& \multicolumn{1}{c}{--} &1.73&1.59\\
\textbf{\textsc{ru}} &1.82&1.54&1.64&1.72&-0.84^*& \multicolumn{1}{c}{--} &0.80^*\\
\textbf{\textsc{tr}} &1.88&0.98^*&1.88&1.70&-1.80&0.58^*& \multicolumn{1}{c}{--} \\
\bottomrule
\end{tabular}
}
\caption{XWEAT T9 effect sizes for cross-lingual embedding spaces. Rows denote the target set language, column the attribute set language.}
\label{tbl:xling_wiki_9}
\end{table}
\end{document}